\documentclass[11pt,a4paper]{article}

\usepackage[acceptedWithA]{tacl2021v1}

\usepackage{times,latexsym}
\usepackage[T1]{fontenc}
\usepackage[utf8]{inputenc}
\usepackage{microtype}
\usepackage{inconsolata}

\usepackage{graphicx}
\usepackage{changepage}
\usepackage{float}
\usepackage{subcaption}
\usepackage{rotating}
\usepackage{makecell}
\usepackage{amsmath}
\usepackage{pdflscape}
\usepackage{multirow}
\usepackage{booktabs}
\usepackage{enumitem}
\usepackage{tabularx}
\usepackage{comment}

\newcommand{\sent}[1]{\textit{#1}}
\newcommand{\sem}[1]{\texttt{#1}}

\newcommand{\textcite}[1]{\citet{#1}}
\newcommand{\parencite}[1]{\citep{#1}}

\title{Language Models Can Resolve Reference Compositionally,\\
But It’s Not Their Native Strength: The Case of the Personal Relation Task}

\author{Bart Evelo \hspace{1em} Meaghan Fowlie \hspace{1em} Denis Paperno 
\vspace{1em} } 
\date{}

\begin{document}
\maketitle

\begin{abstract}
Do neural models, such as Large Language Models, genuinely acquire compositional abilities for interpretation of natural language?
When we talk about semantic interpretation, we can distinguish two complementary aspects: establishing what an expression refers to in the world (which we call the Extensional task) and representing its sense in a structured way (which we call the Intensional task). We evaluate LLMs and humans on both tasks in the setting of the Personal Relation Task \citep{Paperno2022} in which, given a universe of people and their relationships with each other, one is asked to interpret a noun phrase such as \sent{Amber’s parent’s friend}. Here, for the Intensional task, the answer is the formula \sem{friend(parent(amber))}, and for the Extensional task, the person. 
We find that humans and LLMs show opposite strengths: humans perform better on Extensional than Intensional tasks, and LLMs \textit{vice versa}.
Our methodology brings greater nuance to the understanding of compositional abilities in modern machine learning models. 
Our results support the notion that the lack of referential grounding in LLM training is a crucial missing component in mimicking human-like language understanding.  
\end{abstract}
\section{Introduction}
\label{sec:Introduction}

Human languages can have an infinite number of meaningful sentences built from a finite set of elements \cite{Chomsky1957, Montague1970}. This is made possible by the recursive structure of sentences and the principle of \textit{semantic compositionality} \citep{jannsen,carnap1988meaning}, whereby ``the meaning of a whole is a function of the meanings of the parts and of the way they are syntactically combined'' \cite{Partee1995}. 
 Compositional structure has been argued to be not just a theoretical construct, but a property of language data that benefits deep neural networks trained on them \cite{galke2024deep}. 
 
 In semantic theory, the discussion of compositionality is intertwined with the important contrast between two aspects of meaning: sense and reference
\citep{Frege1892-FREBSU}. A noun phrase's \textit{reference} (\textit{Bedeutung}) is the actual entity it denotes in a given world, whereas its \textit{sense} (\textit{Sinn}) is the abstract concept or computational procedure that determines this reference. These are the objects of \textit{Extensional} and \textit{Intensional} semantics, respectively. Two expressions can have the same reference but different senses. If \textit{Amber's friend} and \textit{Bryan's enemy}  both refer to the same person, we can describe their senses using two different formulae \sem{friend(amber)} and \sem{enemy(bryan)} but represent their reference identically as e.g.,\ \sem{christina}.
 
The linguistic fluency demonstrated by Large Language Models (LLMs) is superficially similar to human performance in many aspects \cite[e.g.][]{Yang2024Babbling}. A growing body of research \cite[cf.\ survey][]{sinha2024surveycompositionallearningai} addresses a critical question: do AI models genuinely acquire compositional abilities for interpretation, or do their impressive linguistic outputs result primarily from sophisticated pattern matching at the form level? 

Much of this work on compositional generalization of LLMs relies on variants of semantic parsing, translating natural language into a formalism; COGS \cite{Kim2020} is a representative benchmark. The referential aspect of compositional meaning and its relation to conceptual meaning (sense, which semantic parses can be seen as describing), have not received enough attention.

\begin{figure*}
\vspace{2.7em}
\begin{tabularx}{\textwidth}{|l|l|X|cc|cc|}
\multicolumn{1}{c}{\multirow{2}{*}[4em]{\rotatebox[origin=c]{90}{\parbox{3cm}{\centering\textbf{Complexity}}}}} &
\multicolumn{1}{c}{\multirow{2}{*}[4em]{\rotatebox[origin=c]{90}{\parbox{3cm}{\centering\textbf{Branching}}}}} &
\multicolumn{1}{c}{\multirow{2}{*}[2em]{\textbf{Q}: Who is\dots \rule{0pt}{1.5cm}}}  &
\multicolumn{2}{c}{\textbf{A: Extensional}} &
\multicolumn{2}{c}{\textbf{A: Intensional}} \\
\cline{4-7}
\multicolumn{1}{c}{}&\multicolumn{1}{c}{}&\multicolumn{1}{c|}{}& \textbf{English} & \textbf{Abs.} & \textbf{English} & \textbf{Abs.} \\
\hline
\hline
3 & L & Amber's parent's friend & Felicia & \sem{q} & \sem{friend(parent(amber))} & \sem{x(k(h))}\\
3 & R & the friend of the parent of Amber & Felicia & \sem{q} & \sem{friend(parent(amber))} & \sem{x(k(h))}\\
4 & L & Amber's parent's friend's parent & Chris & \sem{y} & \sem{parent(friend(parent(amber)))} & \sem{k(x(k(h)))}\\
4 & R & the parent of the friend of the parent of Amber & Chris & \sem{y} & \sem{parent(friend(parent(amber)))} & \sem{k(x(k(h)))}\\
\hline
\end{tabularx}
\caption{
Variables in the Personal Relation Task (PRT): Values of complexity, branching (L and R denote Left- and Right branching, respectively), Approach (Extensional vs Intensional), and Representation type (English vs Abstract (Abs.)). Extensional Answers can be found using the model in Figure \ref{fig:universe}, Appendix \ref{sec:app_relation-graph}.}
\label{table:prt-variables}
\end{figure*}

To address this, we adapt 
the \textbf{\textit{Personal Relation Task (PRT)}} \citep{Paperno2022}, in which one is asked to interpret complex noun phrases such as \sent{Felicia's parent's enemy's child}. The complexity of the phrases requires recursive compositional interpretation; it is highly unlikely that a human or an LLM would have previously learned each of such phrases as a holistic unit. 
Taking an example from our version of the PRT, given a question like \sent{Who is the friend of the parent of Amber?}, the reference is the specific person the phrase points to (e.g., Felicia). 
We call this the \textit{\textbf{Extensional Task.}} The sense, in contrast, is the 
conceptual structure of the query, captured in a formal representation like \sem{friend(parent(amber))}, which we dub the \textbf{\textit{Intensional Task}}. This distinction is critical, as a model could correctly produce the formal sense without being able to resolve its concrete reference, or vice versa. 

In ordinary communication, people routinely identify intended referents. While this process requires comprehending the intensional structure of the query, humans perform this analysis implicitly and rarely need to resort to explicit notation for it. 
Therefore, we expected humans to be better at the Extensional task than the Intensional task; this prediction was borne out.

Unlike humans, whose referential abilities are rooted in real-world interaction, LLMs operate only over text. As a result, they  have limited access to grounding beyond linguistic data but excel at translation; therefore, 
we expected LLMs to be better at the Intensional task. 
 This too was borne out, and the difference increased as the complexity of the noun phrases increased. \S\ref{sec:Methodology} gives the full experimental set-up, summarized in Fig.\ \ref{table:prt-variables}, and \S\ref{sec:Results} and \S\ref{sec:Discussion} discuss the full results.
 
 Our contributions include: 
\begin{itemize}[noitemsep, topsep=0pt, leftmargin=*]
    \item a methodology for assessing intensional vs.\ extensional compositional interpretation 
    \item data on human performance on the same tasks in maximally comparable setup to LLMs
    \item evaluation of several large language models and statistical analysis of model performance, highlighting different strengths of LLMs compared to human participants.
\end{itemize}
\section{Related Work}
\label{sec:Background}

When discussing how sense vs.\ reference relate to LLMs, some authors \citep{merrill2021provable,allencarnap} theorize about LLMs learning or implementing extensional vs intensional models from text, while other \citep{bouyamourn2023llms} connect intensionality to paraphrasing, a task which LLMs excel at since the T5 model \cite{raffel2020exploring}. However, when it comes to empirical evaluation of LLMs, intensional and extensional interpretation have not been contrasted explicitly.

Many of the studies in compositional learning are of immediate relevance here \cite{sinha2024surveycompositionallearningai}. Work on compositional abilities of neural models highlights their limitations, e.g.,\ \citet{Lake2018,yao-koller-2022-structural}, but there are also reports on successes \cite{lake2023human,yao-koller-2024-simple,Zhou2022}. 
Existing benchmarks for the evaluation of compositional abilities in AI can be broadly categorized by whether they test for an extensional or an intensional understanding of language. Extensional tasks require a model to map descriptions to referents. SCAN \citep{Lake2018} can be seen as such a task: mapping a command to a concrete sequence of actions. Another example is the realistic compositional instruction following benchmark by \citet{yang2024exploring}. Conversely, intensional tasks, which include COGS \cite{Kim2020}, GeoQUERY \cite{zelle1996learning}, SPIDER \cite{yu2018spider}, and others  \cite{Keysers2020Measuring,zhang-etal-2024-gaining} focus on translating a sentence into a formal, abstract representation, such as a logical formula or a database query, which can be seen as a formalization of sense. While these benchmarks have been crucial for identifying compositional abilities and weaknesses in neural architectures, they typically test for one form of understanding or the other. A key goal of our work is to use a single, unified framework to probe both intensional and extensional interpretation in a comparable way.

Two properties might have systematically affected the difficulty of compositional tasks for language models. First, most existing tasks are intensional, whereby compositionality largely boils down to piecemeal translation from a natural language to a formal one. 
\citet{Shaw2021} and a follow-up study by \citet{Sun2023} found that high scores on SCAN do not guarantee good performance on datasets like GeoQUERY and SPIDER, and vice versa, which they interpret in terms of synthetic vs.\ natural data. However, one could also interpret the contrast in terms of the largely extensional nature of SCAN and intensional nature of other datasets. The relevance of the extensional/intensional distinction, as opposed to natural/synthetic, is further suggested by the fact that in \citeauthor{Sun2023}'s experiment, T5 performed well on all intensional datasets, including synthetic COGS, but not on the extensional SCAN.

Second, benchmarks like COGS reuse elements of the language inside meaning representations for it, e.g.,\ word \textit{cat} translates into logical constant \sem{cat}; neural systems are known to exploit this, relying on mere string copying \cite{zhang2025neural}. However, in semantic theory, forms (e.g.,\ strings) and meanings (e.g.,\ entities) are objects of different nature generally without an inherent relationship to each other. In our experiments, we control for the transparency of elementary meaning representations as the separate \textit{representation} variable.

With notable exceptions such as \citet{lake2019humanfewshotlearningcompositional}, compositional abilities of models are often assessed without an explicit comparison to humans. 
Our work aims to fill these gaps. We design an evaluation for intensional vs.\ extensional compositional representations in comparable conditions; we control for the transparency of elementary meaning representations and provide a comparison with human behavior on the same data.
\section{Experimental Setup}
\label{sec:Methodology}

To evaluate compositional reasoning, we developed a series of experiments based on a modified and extended version of the Personal Relation Task (PRT), originally designed by \textcite{Paperno2022}. This task provides a controlled environment to test semantic compositionality in both Large Language Models (LLMs) and human participants.

\subsection{Task Design and Variables}
\label{sec:task-design}

The core of the task is a self-contained "universe": a graph of 6 nodes (people) connected by edges representing one of four relations: \sem{friend, enemy, parent}, or \sem{child}. The \sem{friend} and \sem{enemy} relations are symmetric, while \sem{parent} and \sem{child} are asymmetric inverses. (See 
Fig.\ \ref{fig:universe} in Appendix \ref{sec:app_relation-graph} for an example universe.) Questions were generated by recursively traversing this graph; edges used in the example were excluded.  To probe different aspects of compositional reasoning, we manipulate four key variables (Fig.\ \ref{table:prt-variables}) as follows:

\paragraph{Approach: Extensional vs. Intensional.} 
The Extensional task is to identify the referent of the given noun phrase, while the Intensional task is to provide a nested functional notation, such as \sem{friend(parent(amber))}.

\paragraph{Representation: English vs. Abstract.} The relationship between people and their names is arbitrary, as is the relationship between terms like \sent{child} and the idea of being someone's child. To emulate this property of language, we also designed an \textit{\textbf{Abstract}} version of each task, in which the predicates and individuals are mapped to arbitrary variables (e.g., \sent{Amber} = \sem{h}, \sent{friend} = \sem{x}). 
The \textit{\textbf{English}} version uses standard names and relations (e.g., \sent{Amber} = \sem{Amber}, \sent{friend} = \sem{friend}). 

This abstract format was introduced for two additional reasons. First, it acts as a control for pretraining data contamination; by using novel symbols, we can be confident that LLMs have not previously encountered this specific task format in textbook examples or other training materials. Second, it increases the difficulty of the intensional task. We hypothesized that the English intensional task might otherwise be solvable through superficial string manipulation rather than true structural understanding. Requiring models to first map abstract symbols to their meanings before constructing a formula forces a more robust form of compositional analysis.

The English Intensional task does not require the universe at all; the solution just involves rearranging the symbols in the natural language query and adding brackets. The Abstract Intensional task on the other hand requires lookup in the universe for referents for predicates and names. While everything an LLM does is, in a sense, string manipulation, the Abstract variant is less straightforward.

\paragraph{Branching: Left vs. Right.} 
This variable refers to the syntactic structure of the query. Left-branching questions used the genitive \sent{'s} marker (e.g., \sent{Who is Amber's parent's friend?}). Right-branching questions used prepositional phrases (e.g., \sent{Who is the friend of the parent of Amber?}).

\paragraph{Complexity.} 
The complexity of a question is defined as the number of relations involved plus one. LLMs were tested on complexities 3, 4, 5, and 6, while the human experiment was limited to complexity 3 due to budgetary constraints. 
 
\subsection{Stimuli and Procedure}
LLM accuracy was estimated based on a single run. Each stimulus presented to both LLMs and humans followed a standardized four-part structure: (1) a brief task introduction, (2) the complete dataset of relations, (3) a worked out example with a step-by-step solution, and (4) the question. The full dataset, including the generated universes and questions, is publicly available for reproducibility in our GitHub repository\footnote{
\url{https://github.com/avatars16/Personal-Relation-Task-on-LLMs}
}. 
A simplified example (with a four-person universe instead of the six-person universe used in our experiments) is in Fig.\ \ref{fig:prompt}. The four main conditions' prompts differ on their universe representation and worked example. Universe items \sent{the enemy of Amber} (right-branching) and \sent{Amber's enemy} (left-branching) have the following representations:

\begin{tabular}{ll}
    English Extensional & Dana\\
    English Intentional & \sem{enemy(Amber)} \\
    Abstract Extensional & $s$ \\
    Abstract Intensional & \sem{w(h)} \\
\end{tabular}

\begin{figure}[tb]
\fbox{
\centering
\begin{minipage}{0.45\textwidth}
Imagine there are four people: Amber, Bryan, Christina and Dana. They have the following relationships to each other:\\

\begin{tabular}{ll}
the enemy of Amber & = Dana 
\\
the friend of Amber & = Christina \\
the enemy of Bryan & = Christina 
\\ the friend of Bryan & = Dana \\
the enemy of Christina & = Bryan 
\\ the friend of Christina & = Amber \\
the enemy of Dana & = Amber 
\\ the friend of Dana & = Bryan 
\end{tabular}

\vspace{1em} 

\noindent Here is how to figure out the answer to the following question:

\vspace{0.5em} 

\sent{Who is the enemy of the friend of Amber?}

\vspace{0.5em} 

Here are the steps to arrive at the answer:
\begin{enumerate}[noitemsep, topsep=0pt]
    \item the friend of Amber = Christina
    \item the enemy of Christina = Bryan
\end{enumerate}

\vspace{0.5em} 

So the answer is Bryan
\vspace{0.5em} 

Now answer the following question:
\\
Who is the friend of the enemy of Amber?
\end{minipage}
}
\caption{Illustrative example of the experimental stimuli. 
This simplified version uses a four-person universe (instead of six) and only includes the \textit{friend} and \textit{enemy} relations (excluding \textit{parent}/\textit{child}). 
The example shown here is Complexity 3, Extensional, English, Right-branching condition. 
Otherwise, the simplified example follows the same standardized four-part structure as full stimuli: a brief introduction, the dataset of relations, a worked example, and an actual test question.}
\label{fig:prompt}
\end{figure}

\subsubsection{Experiment 1: LLM Evaluation}
We evaluated the five LLMs listed in Table \ref{table:model-comparison-condensed}. We included recently developed Large Reasoning Models \citep[][LRMs]{guo2025deepseek}, which  
attempt to achieve more robust reasoning by internalizing a ``thought" process, usually achieved via reinforcement learning. 
Open-weight models were run in under 21 hours on H100, including debugging. 
For each model, we generated 1,280 prompts ($40 * \text{Approach} (2) * \text{Representation}(2) * \text{Branching}(2) * \text{Complexity}(4)$)  covering every combination of experimental variables. To standardize the output for automated parsing, the instruction \textit{"Please answer the question in the same format as the example. End with 'So the answer is [answer]'."} was appended to every prompt.

\begin{table*}[t]
\centering
\begin{tabular}{llccr}
\toprule
\textbf{Model Name} & \textbf{Model Alias} & \textbf{Reasoning} & \textbf{Open} & \textbf{Reference} \\
\midrule
OpenAI GPT 4.1 & gpt-4.1 & No & No & \citealt{gpt41}\\
OpenAI o3mini & o3-mini & Yes & No & \citealt{openai2025o3mini}\\
Qwen2.5-32B & qwen-2.5 & No & Yes & \citealt{qwen2.5}\\
DeepSeek-R1-Distill-Qwen-32B & deepseek-distill & Yes & Yes & \citealt{DeepSeek-AI2025} \\
Llama-3.3-70B-Instruct & llama-3.3 & No & Yes & \citealt{meta2024llama3.3}\\
\bottomrule
\end{tabular}
\caption{Selected Large Language Models (LLMs).}
\label{table:model-comparison-condensed}
\end{table*}

\subsubsection{Experiment 2: Human Evaluation}
Anonymized data was collected from 40 participants via the crowd-sourcing platform (\url{www.prolific.com}\nocite{prolific}). The final analysis used a sample of 32 after excluding participants who did not achieve 100\% accuracy on four control questions 
consisting of a single relationship (e.g.,\ \textit{Amber's friend}). 

The experiment was implemented in PCIbex \parencite{zehr2018penncontroller}. Participants completed 8 experimental trials  and 4 control trials. Due to budgetary constraints, humans were only tested on complexity 3. For Extensional questions, participants selected an answer from an alphabetized dropdown menu; for Intensional questions, they typed their response into a text field.

\subsection{Analysis}
The primary performance \textbf{metric} 
measures exact match with the correct answer, except a wrong number of closing brackets in intensional answers is not penalized. This allows the analysis to focus on semantic understanding over syntactic precision. To statistically evaluate the factors influencing accuracy, we fitted a series of Generalized Linear Mixed Models (GLMMs) with a binomial family using the \texttt{lme4} package \parencite{Rlme4} in R \parencite{RCoreTeam}. For more details on the statistical analysis procedure, see Appendix \ref{sec:app_stat-analysis-model-selection}.

\subsection{Hypotheses}
\label{sec:hypotheses}

To summarize, we had the following hypotheses about the effects of experimental variables for LLMs and human participants, all of which are supported by our experiments:

\begin{enumerate}[noitemsep, topsep=0pt, leftmargin=*]
    \item LLMs will be better at Intensional than Extensional tasks.\label{item:hyp-LLM-I-E}
    \item Conversely, humans will be better at Extensional than Intensional. \label{item:hyp-h-I-E}
    \item Both humans and LLMs will be better at English than Abstract.\label{item:hyp-A-E}
    \item Unlike older models, LLMs will not show a difference between Left- and Right-branching.\label{item:hyp-LR}
    \item LLMs' performance will decay as complexity goes up.\label{item:hyp-compl}
\end{enumerate}
\section{Results}
\label{sec:Results}
This section presents the comparative performance of Large Language Models (LLMs) and humans on the Personal Relation Task. We analyze the main effects of task approach (Intensional/Extensional), representation (English/Abstract), and complexity (3-6), supported by descriptive statistics and Generalized Linear Mixed Models (GLMMs). Details of the statistical models are provided in Appendix \ref{sec:app_stat-analysis-model-selection} 
and a full table of results is in Appendix \ref{sec:app_accuracies-table}. Across all conditions, humans reached an accuracy of 76.95\%, LLMs reached 87.4\%.

\subsection{Effect of Task Approach: Intensional vs. Extensional}

\begin{figure}[ht]
\centering
\includegraphics[width=\columnwidth, keepaspectratio]{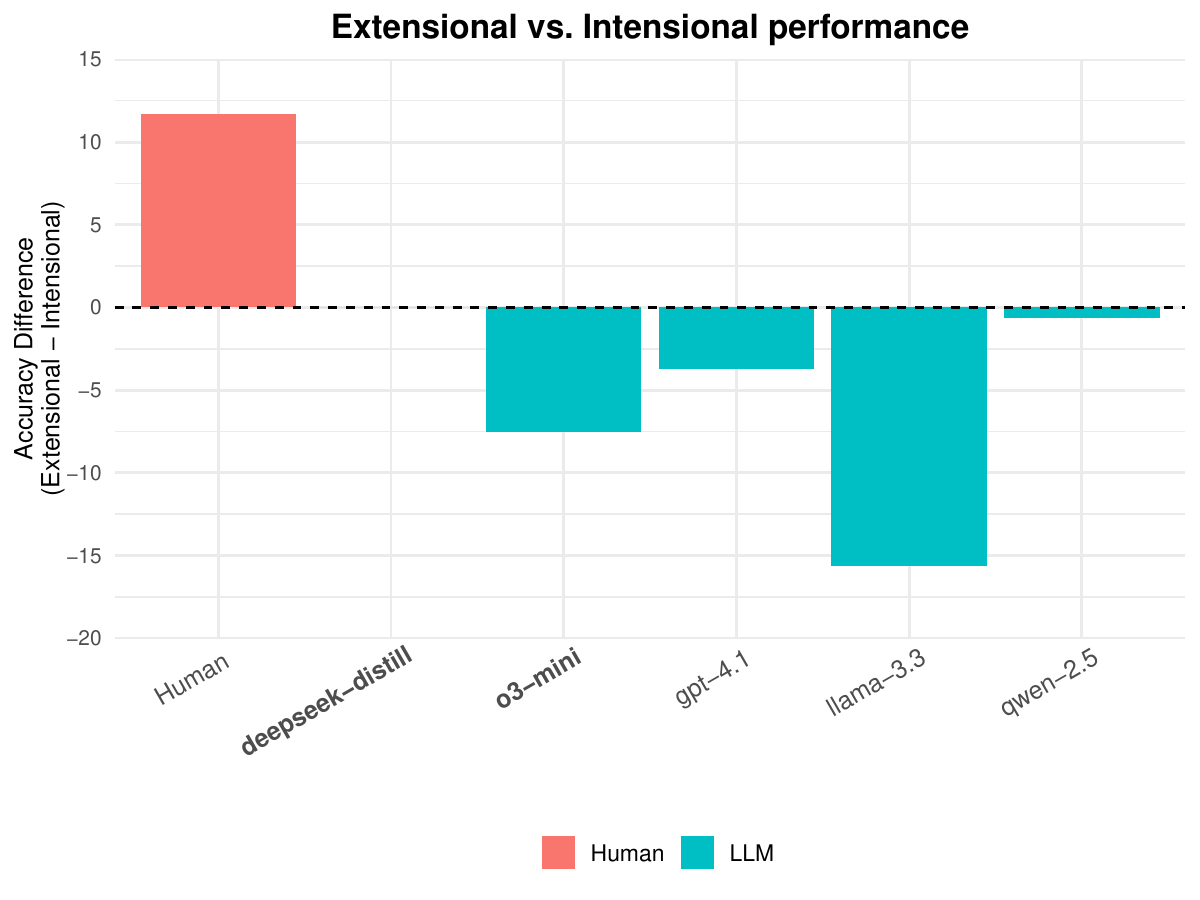}
\caption{Extensional vs. Intensional performance for Humans and LLMs (complexity 3). Reasoning models are printed in bold. The difference is calculated as Extensional accuracy $-$ Intensional accuracy}
\label{fig:diff-extensional-intensional}
\end{figure}

\begin{figure}[ht]
\centering
\includegraphics[width=\columnwidth, keepaspectratio]{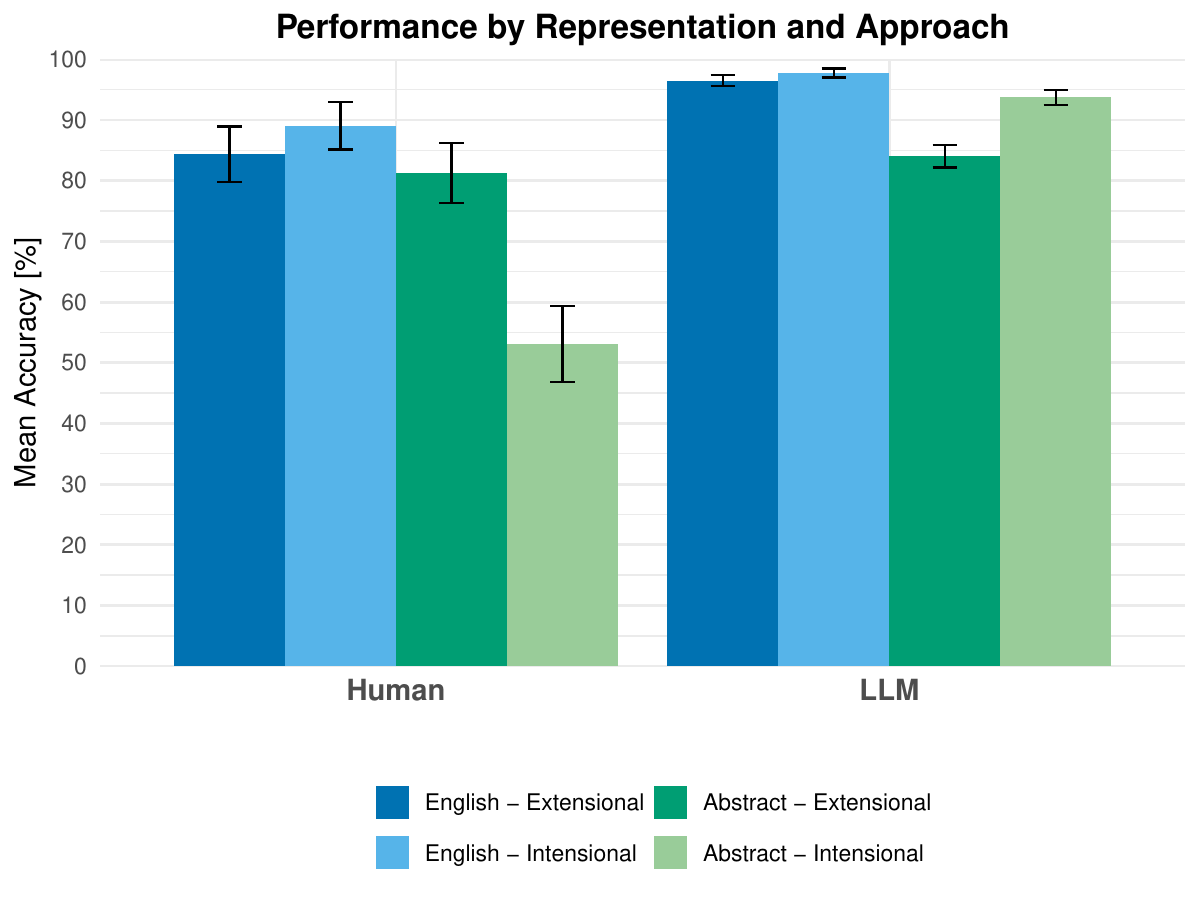}
\caption{Mean accuracy (\%) for Humans and Large Language Models (LLMs) by Representation (English vs. Abstract) and Approach (Extensional vs. Intensional) at complexity 3. The abstract cases (green bars) demonstrate an inverse performance pattern between humans and LLMs: humans found the Extensional approach (dark green) easier than Intensional (light green), while LLMs found Intensional (light green) easier than Extensional (dark green). This divergence becomes clearer at higher complexities.
}
\label{fig:represenation-approach-3}
\end{figure}

Our analysis reveals a significant interaction between the participant group (Human vs.\ LLM) and the task approach, as illustrated in Figure~\ref{fig:diff-extensional-intensional} and \ref{fig:represenation-approach-3}. Humans performed better on the Extensional task (82.8\% accuracy) than the Intensional task (71.1\%). The GLMM for human performance showed a significant negative effect for the Intensional approach (\(\beta = -1.08, p < 0.01\)).

\begin{figure}[htbp]
\centering
\includegraphics[width=\columnwidth]{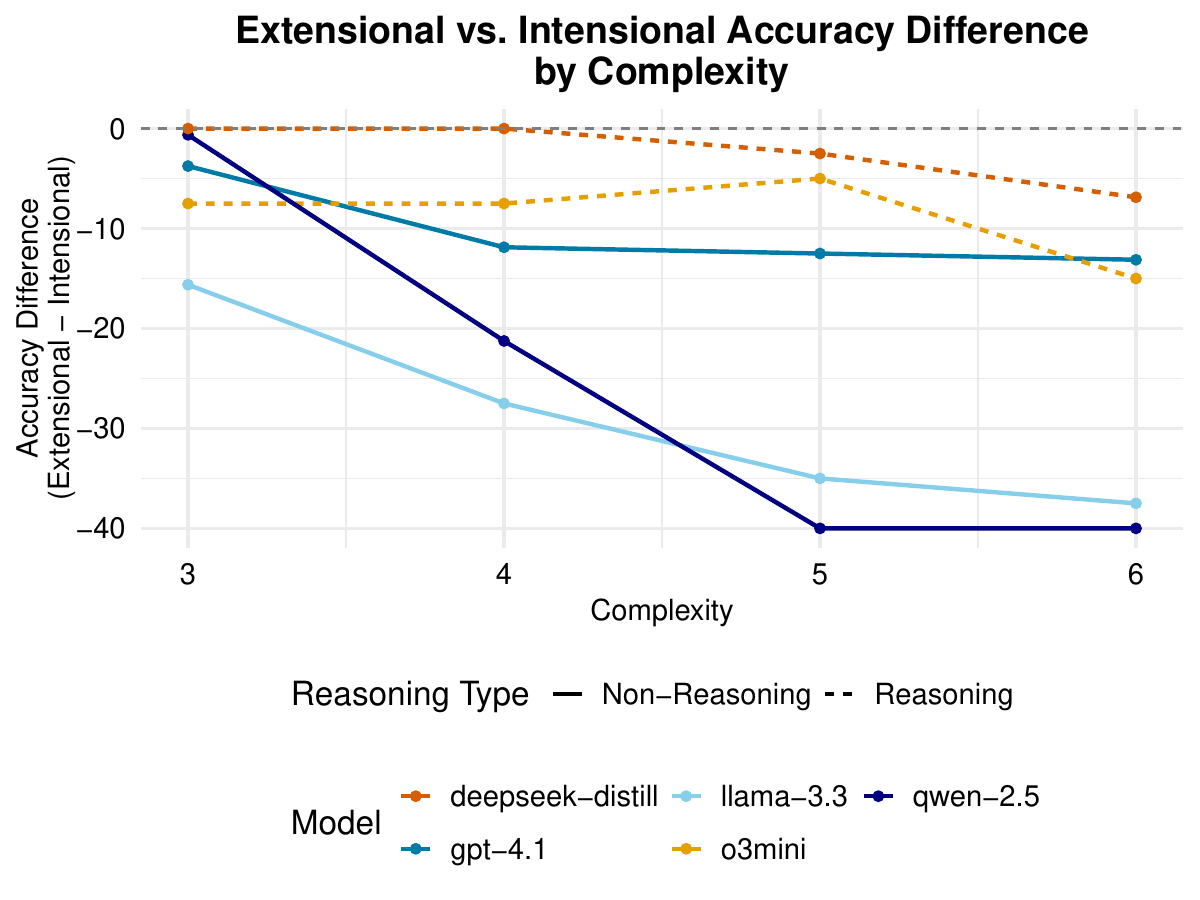}
\caption{Extensional vs. Intensional performance for different complexities. Data shown is from the LLMs tested. The difference is calculated by following formula: Extensional accuracy $-$ intensional accuracy}
\label{fig:diff-extensional-intensional-complexity}
\end{figure}

Conversely, LLMs achieved higher accuracy on the Intensional task (95.0\%) than on the Extensional task (79.8\%). This was reflected in the LLM GLMM as a large positive effect for the Intensional approach (\(\beta = 1.29, p < 0.001\)). This contrast between the human and LLM pattern can be seen in Fig.\ \ref{fig:represenation-approach-3}. As seen in Fig.\ \ref{fig:diff-extensional-intensional-complexity}, for non-reasoning models, this performance gap widens significantly as complexity increases, whereas reasoning-enhanced models maintain a much flatter and more consistent performance difference across all complexity levels. 

A GLMM comparing Humans to LLMs analyzing complexity-3 tasks confirmed this difference. It showed a negative effect of intensionallity for humans \(\beta = -2.05, p<0.01 \) while having a strong positive interaction for LLMs with intensionallity 
\((\beta = 3.17, p < 0.001)\).

\subsection{Effect of  Task Representation Type
}

Both LLMs and humans found tasks with abstract representations more difficult than those presented in English. Human accuracy dropped from 85.0\% in the English condition to 74.0\% in the abstract condition. The human-specific GLMM showed a significant positive effect for the English representation (\(\beta = 1.29, p < 0.001\)).

Similarly, the LLMs' average accuracy was higher for English tasks (95.1\%) than for abstract ones (79.7\%), with a significant positive effect for English representation in the LLM GLMM (\(\beta = 0.84, p < 0.05\)). The combined GLMM did not find a significant interaction between participant group and representation 
\((p = 0.852)\). The lowest accuracy for LLMs (66.1\%) was observed in the abstract-extensional condition.

\subsection{Effect of Task Complexity on LLMs}

For LLMs, task complexity had a significant negative impact on performance. As shown in Table~\ref{tab:llm-complexity}, the average accuracy across all models decreased from 93.0\% at complexity 3 to 82.4\% at complexity 6. The LLM GLMM confirmed this trend with a significant negative main effect for complexity (\(\beta = -0.43, p < 0.001\)). Individual models showed different degradation curves, with \texttt{gpt-4.1} being the most robust. (Humans were tested only on complexity 3.)

\begin{table}[h]

\centering
\small

\begin{tabular}{lcccc|c}

\toprule

\textit{Model} & \textit{3} & \textit{4} & \textit{5} & \textit{6} & Av.\\

\midrule

qwen-2.5        & 92.2 & 82.5 & 79.4 & 76.9 & 82.8\\
deepseek-distill & 91.9 & \textbf{95.0} & 90.0 & 89.1 & 91.5 \\
gpt-4.1          & \textbf{98.1} & 94.1 & \textbf{93.1} & \textbf{92.8} & \textbf{94.5}\\
llama-3.3        & 88.4 & 75.6 & 74.4 & 68.8 & 76.8 \\
o3-mini          & 94.4 & 93.8 & \textbf{93.1} & 84.4 & 91.4 \\
\hline
Average & 93.0 & 88.2 & 86.0 & 82.4 & 87.4\\
\bottomrule

\end{tabular}

\caption{Model accuracy across different question complexities for LLMs, aggregated across all versions of the PRT.}

\label{tab:llm-complexity}

\end{table}

\subsection{Branching Effects}

Branching direction did not yield a significant main effect in the human GLMM (\(p = 0.326\)) or show a consistent pattern across LLMs, though the (non-reasoning) \textit{qwen-2.5} did struggle slightly more with right-branching (\(\beta = -0.94, p < 0.05\)), especially in the Extensional task (\(\beta = 2.00, p < 0.05\)).

\subsection{Focused Analysis: LLM vs.\ LRM} To isolate the impact of reasoning-specific fine-tuning, we compared the standard \textit{qwen-2.5} model with its reasoning-enhanced \textit{deepseek-distill} variant. 
The analysis revealed several significant interactions. A strong positive interaction was found between the standard model (\textit{qwen-2.5}) and the Intensional approach (\(\beta = -2.27, p < 0.001\)), showing the performance gap between Extensional and Intensional tasks was larger for the standard model. 
Furthermore, significant interactions were identified between model type and both complexity (\(\beta = 0.31, p < 0.05\)) and branching direction (\(\beta = 0.81, p < 0.01\)), indicating that the reasoning model compensates for the weaknesses of the standard model when complexity increases (as seen in Fig.\ \ref{fig:diff-extensional-intensional-complexity}), and in right-branching noun phrases.

\section{Discussion}
\label{sec:Discussion}

The broad question of interest in this work is the nature of the ``reasoning'' that LLMs perform when asked to semantically interpret natural language. The particular task is to interpret noun phrases in the context of a universe of people and their relationships to one another.

LLMs are highly proficient at text-based tasks, particularly translation
; one of successes of GPT-3 was achieving SOTA BLEU in translation from French and German to English \cite{brown2020language}.
The intensional interpretation task strongly resembles translation; for English noun phrases, it can often be reduced to a syntactic rearrangement of elements. For example, converting \sent{Amber's parent's friend} to \sem{friend(parent(Amber))} is a structural transformation. Given their proficiency in similar tasks, we predicted that LLMs would find this intensional task of rearranging a noun phrase into a formula easier than the extensional task of navigating the complex universe to find the correct referent (Hypothesis \ref{item:hyp-LLM-I-E}). This hypothesis is related to Bender and Koller's (\citeyear{bender2020climbing}) argument that the lack of grounding in the world outside text limits semantic capabilities of language models. Even though the experimental setup presents a textual representation of a ``world'' of individuals and their relationships, models trained on text may be less predisposed to use such world information while excelling at translation-like tasks, while humans, whose language use is fundamentally grounded in interactions with objects in the outside world, would exhibit the opposite tendency (Hypothesis \ref{item:hyp-h-I-E}).

However, this contrast requires a nuanced interpretation. Our results should not be taken to mean that humans lack compositional intensional understanding. On the contrary, to succeed at the Extensional task, participants must correctly parse the recursive structure of phrases like \textit{Amber’s parent’s friend}. Humans process this structure implicitly; the difficulty arises only when they are asked to make this structure explicit in a formal notation, which is a translation step foreign to natural communication. LLMs show the inverse profile. Their training objective, which focuses on surface-form prediction, makes them highly proficient at the explicit ``translation" required for the Intensional task. Yet, they lack the direct referential grounding acquired through interaction. While humans learn reference explicitly, LLMs likely acquire referential capabilities only implicitly and indirectly during pre-training, to the extent that tracking reference aids in the core objective of next-token prediction.

Hypotheses \ref{item:hyp-LLM-I-E} and \ref{item:hyp-h-I-E} are well supported by the results. On average, LLMs performed 15.2\% worse on the Extensional tasks than the Intensional tasks. Conversely, humans performed 11.7\% better on Extensional than Intensional (Fig.\ \ref{fig:diff-extensional-intensional}). This interaction is statistically significant ($\beta = 3.17, p< 0.001$).

While LLMs demonstrated strong performance on the tasks, maxing out near 100\% and averaging 87.4\% accuracy overall, they show a markedly different pattern from humans. 
English Intensional is a particularly easy translation task, with an English vocabulary but a different syntax, and here LLMs score highest (95\%). In the full LLM experiment, both Intensional ($\beta=1.29, p<0.001$) and English ($\beta=0.84, p<0.05$) made the task significantly easier for the models. Similarly, Abstract Extensional is the farthest from being a straightforward translation operation 
-- the names must be translated into variables and the universe much be navigated -- and this task is by far the hardest for all LLMs, with an average score of 66\%, versus an average of 99.4\% for the other three conditions in the full LLM experiment. 

As expected (Hypothesis \ref{item:hyp-A-E}), both humans and LLMs performed better on English than Abstract (11\% and 15\% better respectively). That said, LLMs performed quite well on the Abstract tasks (79\%). The arbitrariness of the sign is not too much of a hindrance.

For humans, we found the expected results of Extensional (identifying the person, 83\%) being easier than Intensional (creating a formula, 71\%) ($\beta=-1.08, p<0.01$). Unexpectedly, our participants did not find the English Intensional task significantly harder than the English Extensional task\footnote{89\% vs 84\% accuracy. The best statistical model did not include the interaction, so this difference could be statistical noise. Additionally, one pilot showed the opposite trend, with English Extensional easier than English Intensional.}. 
Switching from English to Abstract made the Intensional task extremely hard, however, with participants scoring only 53\% when they needed to write formulae with variables such as \sem{x(y(z))}. 
As expected, English was overall much (29\%) easier for people than Abstract ($\beta=1.29, p<0.001$). 
This supports the idea that humans excel when they can simply tap into their linguistic abilities.

\paragraph{Reasoning Models:} 

While a broad comparison of reasoning versus non-reasoning models was inconclusive due to model heterogeneity, Figure \ref{fig:diff-extensional-intensional-complexity} demonstrates that the performance gap between difficult extensional and easier intensional tasks remains relatively small and stable for reasoning models as complexity increases, unlike standard models where the gap widens considerably. Furthermore, a focused analysis of our single minimal pair —\textit{qwen-2.5} and its reasoning-distilled variant \textit{deepseek-distill} — offers a meaningful insight. This targeted comparison revealed that the reasoning model significantly narrows the performance gap between the difficult referential (Extensional) task and the easier symbolic (Intensional) task.  Though a conclusion based on a single minimal pair of models, this finding supports the idea that the extensional reasoning weakness we identified in LLMs is a specific, addressable challenge rather than an inherent architectural limitation.

\paragraph{Branching (Hypothesis \ref{item:hyp-LR}):} We hypothesised that unlike for earlier results for recurrent neural networks \citep{Paperno2022}, modern Transformer LLMs would not show a branching bias, as the processing is not strictly sequential. This is indeed what we found, with no main effect of branching and only slight and hard-to-interpret interaction effects for one model (\textit{qwen-2.5}).

\paragraph{Complexity (Hypothesis \ref{item:hyp-compl}):} 
LLMs show the expected decay in performance as complexity increases ($\beta=-0.43, p<0.0001$). Task complexity correlates with both recursion depth of the English expression and the number of steps -- i.e.\ opportunities for mistakes -- required to complete the task, so this can be an issue of language, task complexity, or both.

\section{Conclusion}
In this work, we formalized the task of representing compositional meanings inspired by formal semantics and analytic philosophy. Both the intensional and the extensional versions of the task follow this tradition. The abstract version further follows the spirit of the analytical tradition, although in practice formal semantic representations often resemble our English version of the task, motivated by human readability.  The Personal Relations Task allowed us to test all these kinds of semantic representations in a way that is accessible both to laypeople (our experimental participants) and to language models, so we could compare the two.

In contrast to previous generations of neural models, the LLMs in our study show a remarkable progress. Even the weakest LLMs we tested give accurate responses most of the time, unlike smaller models tested previously \cite{bezema2019,monster2021,dewolf2023}. Compared to recurrent architectures tested by  \citet{Paperno2022}, LLMs do not suffer from systematic structural biases, and don't require extensive task-specific training for reasonable accuracy, as in Paperno's or Yao et al.'s \citeyear{yao-koller-2024-simple} experiments.   
And while the extensional version of the task remains relatively difficult for LLMs, adding reasoning seems to reduce the gap. 

We deliberately made semantic compositionality  accessible to language models by representing meaning through text;  sticking to English words makes the task even easier. More realistic interpretation of referring expressions will require going beyond textual inputs and outputs.

In a possible future version of the task, not only could extensions (referential meanings) be presented e.g.,\ visually, but also the outputs could include not just symbolic labels but pointers to a perceptual space, e.g.,\  to image regions.

Another interesting future direction involves mechanistic interpretability. In the current paper, we treated compositional meaning representation as an explicit task. One may wonder whether model internals also include similar compositional mechanisms, as suggested in the literature \cite{biology2025,yao-koller-2024-simple}, and to what extent such mechanisms are critical for the downstream performance of the model.

In sum, LLMs can be very good at different versions of the compositional Personal Relations Tasks. However, they do not behave like humans, who are better at identifying referents, while LLMs are better at writing formulae representing meanings. This parallels findings in the planning domain, where GPT-4 was found to be better at representations of planning problems than at solving them directly \cite{liu2023llm+}. Whatever LLMs are doing -- arguably, leaning on their translation capabilities -- it is not what humans do; LLMs might be achieving similar or better outcomes through a different underlying process. Arbitrary reference still provides a noticeable challenge even for frontier models, supporting the argument \citep{bender2020climbing,boleda2020distributional,xu2025large} that referential grounding may be an important missing component to integrate in AI systems.

\section*{Limitations}
Pursuing the goal of comparing compositional interpretation in human participants vs.\ LLMs, we created a setup for both that was as comparable as possible, but also somewhat artificial.
Results of the human experiment are therefore context specific. Prolific participants performing the task of referential interpretation from detailed textual instructions under time pressure should not be taken as representative of human behavior in all contexts. Due to budget limitations, we were unable to test humans on higher complexities, and overall, the amount of human data is lower than ideal, which we suspect is a contributor to the amount of statistical noise in this experiment.

LLMs, in their turn, were evaluated in favorable conditions, with an explicit textual encoding of the compositional task in a one-shot setting with a fully worked out reasoning example. In more natural settings, LLM performance in tasks involving compositional reference resolution may be worse. 

We leave a broader comparison of basic instruction-tuned vs.\ reasoning enhanced models for future research. Here, we report results for only one minimal pair.

As many empirical studies of LLMs, our findings may be limited to the kinds of current models that we examine. Future models may exhibit somewhat different properties. We also note that a single prompt format was chosen prior to all experiments; different prompt formats were not explored.

\section*{Ethical Considerations}
The human data collection protocol has been approved by the Faculty Ethics Assessment Committee -- Humanities at Utrecht University. 
Participants were provided an information form at 
\url{https://fetc.hum.uu.nl/proposals/attachments/2416/download/321/}. 
Participants received \pounds2.50 for completing the questionnaire. Compensation level was estimated on the basis of the hourly rate of \pounds 10 and based on experiment duration confirmed in a pilot study.

We used ChatGPT when revising the text of this paper, mainly to paraphrase individual sentences.

\section*{Acknowledgements}
The research has been conducted with the support of the Institute for Language Sciences and the Center for Digital Humanities at Utrecht University. We thank Raffaella Bernardi, Kees van Deemter, Pablo Mosteiro, action editor Nathan Schneider, anonymous reviewers, and the audience of IWCS 2025 for helpful feedback on earlier versions of this work.

\bibliography{master}
\bibliographystyle{acl_natbib}

\appendix
\newpage

\twocolumn
\section{Relation graph}
\label{sec:app_relation-graph}

\begin{figure}[htbp]
    \centering
    \fbox{%
        \includegraphics[width=\columnwidth,
        keepaspectratio,trim=5pt 4pt 4pt 5pt, 
        ]{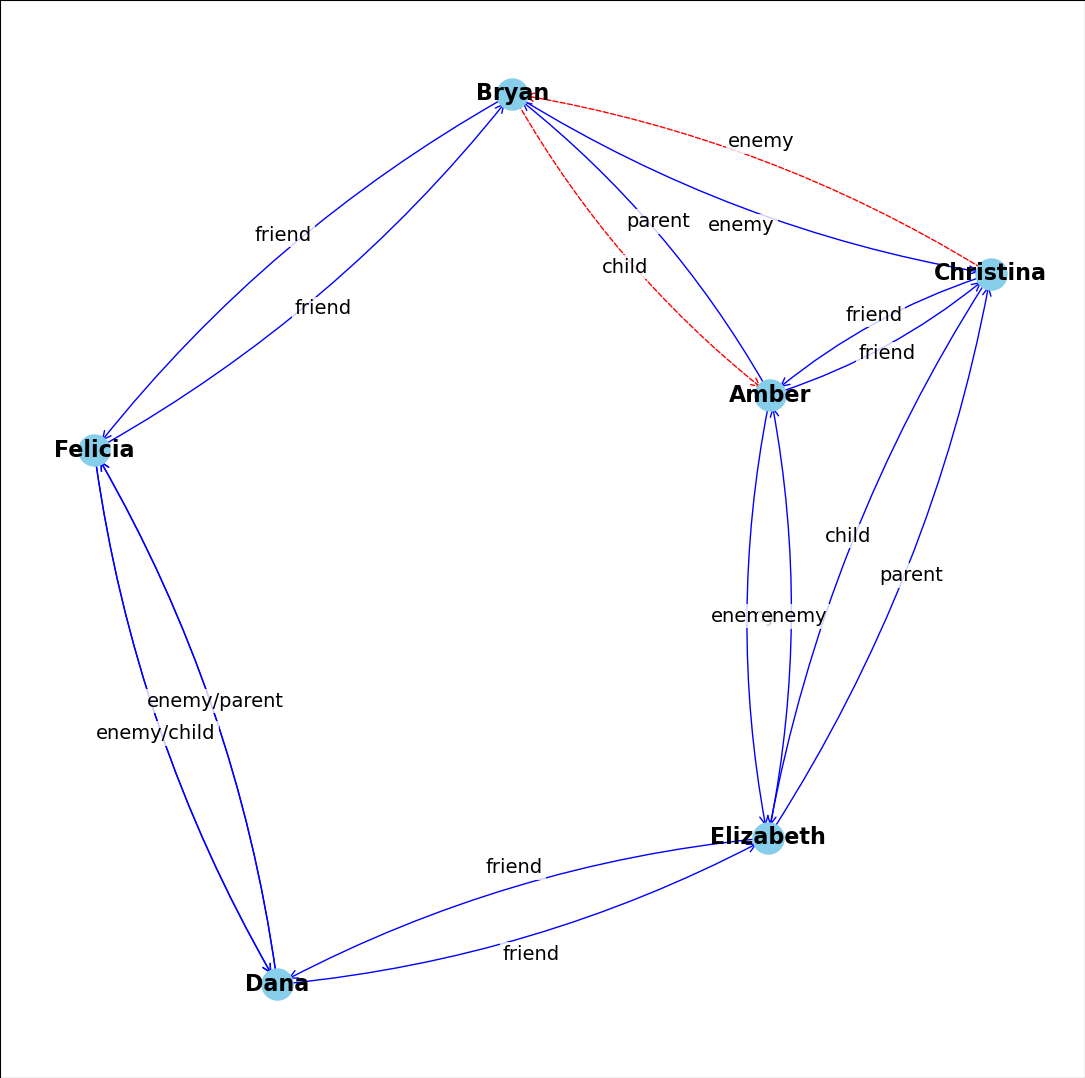}%
    }
    \caption{Graph of persons and their relations to each other used to sample paths for in the stimuli. The paths used in the example are given in red. They were excluded from sampling in the questions.}
    \label{fig:universe}
\end{figure}

\section{Statistical Analysis and Model Selection}
\label{sec:app_stat-analysis-model-selection}

To evaluate the factors influencing accuracy across our experiments, we employed Generalized Linear Mixed Models (GLMMs) with a binomial family and a logit link function. These models were implemented using the lme4 package \parencite{Rlme4} in R \parencite{RCoreTeam}. The primary performance metric, Correct\_Forgiving, was used as the binary outcome, which tolerates minor syntactic deviations (e.g., mismatched closing brackets in intensional answers) to focus on semantic understanding rather than strict output format adherence.

For each experiment (LLM performance, Human Experiment, as well as the combined Human-LLM comparison), a systematic model selection process was followed:

\begin{enumerate}

\item Baseline Model: We began by fitting a baseline GLMM including all main effects of the experimental variables relevant to the specific dataset (e.g., Model, Complexity, Branching, Approach, Representation for LLM analysis; Experiment, Branching, Approach, Representation for the combined analysis). Random intercepts were included for ParticipantId (in human experiments and combined models) and PathId (in LLM experiments, and explored for humans).

\item Interaction Terms: We incrementally added interaction terms, starting with two-way interactions, then three-way, and finally four-way interactions where appropriate. The inclusion of interaction terms was guided by initial descriptive analyses and theoretical expectations about how variables might jointly influence performance.

\item Model Comparison: Likelihood Ratio Tests (LRTs) were used to compare nested models. A more complex model was selected over a simpler one if it provided a statistically significant improvement in fit (typically with \(p < 0.05\)). Akaike Information Criterion (AIC) and Bayesian Information Criterion (BIC) were also considered to balance model fit with parsimony.

\item Diagnostic Checks: After selecting the final model for each analysis, DHARMa diagnostic plots \parencite{DHARMa} were generated to assess the model's distributional assumptions (residuals vs. predicted values, dispersion, and inflation). These checks ensured that the chosen GLMM structure adequately described the data and its error characteristics, as detailed in Appendix B.

\end{enumerate}

For the specific GLMM on data from just LLMs
, the reference model for the Model factor was o3-mini due to its consistent high performance. The gpt-4.1 model was excluded from this GLMM analysis due to its near-ceiling performance, which resulted in limited data variability and convergence issues for the statistical model.

For the combined GLMM comparing Human and LLM performance
, the data for LLMs was subsetted to only include complexity level 3, matching the human experiment data to ensure a fair comparison. The reference levels for this model were Experiment (Human), Branching (Left), Approach (Extensional), and Representation (Abstract).

This rigorous approach to model selection ensures that the reported fixed effects and interaction terms provide the most statistically robust explanation for the observed patterns in compositional reasoning performance.
\clearpage
\onecolumn

\begin{landscape}
\section{Accuracies table}
\label{sec:app_accuracies-table}

\begin{table*}[ht]
\centering
\small
\begin{tabular}{c|l|cccccccc}
\hline
Complexity & Model & Ext.\ Abstract L & Ext.\ Abstract R & Ext.\ English L & Ext.\ English R & Int.\ Abstract L & Int.\ Abstract R & Int.\ English L & Int.\ English R \\
\hline
3 & \textit{Human} & 78.1\% & 84.4\% & 84.4\% & 84.4\% & 50.0\% & 56.3\% & 87.5\% & 90.6\% \\
3 & deepseek-distill & 80.0\% & 90.0\% & 100.0\% & 97.5\% & 77.5\% & 90.0\% & 100.0\% & 100.0\% \\
3 & o3-mini & 95.0\% & 80.0\% & 92.5\% & 95.0\% & 97.5\% & 95.0\% & 100.0\% & 100.0\% \\
3 & llama-3.3 & 70.0\% & 65.0\% & 87.5\% & 100.0\% & 95.0\% & 92.5\% & 97.5\% & 100.0\% \\
3 & gpt-4.1 & 87.5\% & 97.5\% & 100.0\% & 100.0\% & 100.0\% & 100.0\% & 100.0\% & 100.0\% \\
3 & qwen-2.5 & 97.5\% & 77.5\% & 95.0\% & 97.5\% & 92.5\% & 97.5\% & 100.0\% & 80.0\% \\
\hline
4 & deepseek-distill & 87.5\% & 92.5\% & 100.0\% & 100.0\% & 87.5\% & 92.5\% & 100.0\% & 100.0\% \\
4 & o3-mini & 77.5\% & 90.0\% & 95.0\% & 97.5\% & 92.5\% & 100.0\% & 100.0\% & 97.5\% \\
4 & llama-3.3 & 47.5\% & 30.0\% & 70.0\% & 100.0\% & 92.5\% & 80.0\% & 87.5\% & 97.5\% \\
4 & gpt-4.1 & 75.0\% & 77.5\% & 100.0\% & 100.0\% & 100.0\% & 100.0\% & 100.0\% & 100.0\% \\
4 & qwen-2.5 & 55.0\% & 35.0\% & 97.5\% & 100.0\% & 95.0\% & 100.0\% & 77.5\% & 100.0\% \\
\hline
5 & deepseek-distill & 77.5\% & 82.5\% & 97.5\% & 97.5\% & 90.0\% & 82.5\% & 97.5\% & 95.0\% \\
5 & o3-mini & 87.5\% & 90.0\% & 95.0\% & 90.0\% & 100.0\% & 87.5\% & 97.5\% & 97.5\% \\
5 & llama-3.3 & 30.0\% & 22.5\% & 90.0\% & 85.0\% & 90.0\% & 85.0\% & 100.0\% & 92.5\% \\
5 & gpt-4.1 & 70.0\% & 80.0\% & 100.0\% & 97.5\% & 100.0\% & 100.0\% & 100.0\% & 97.5\% \\
5 & qwen-2.5 & 25.0\% & 15.0\% & 100.0\% & 97.5\% & 100.0\% & 100.0\% & 100.0\% & 97.5\% \\
\hline
6 & \textbf{deepseek-distill} & 62.5\% & 92.5\% & 92.5\% & 95.0\% & 90.0\% & 85.0\% & 100.0\% & 95.0\% \\
6 & \textbf{o3-mini} & 65.0\% & 75.0\% & 80.0\% & 87.5\% & 85.0\% & 97.5\% & 92.5\% & 92.5\% \\
6 & llama-3.3 & 37.5\% & 27.5\% & 70.0\% & 65.0\% & 85.0\% & 80.0\% & 95.0\% & 90.0\% \\
6 & gpt-4.1 & 75.0\% & 72.5\% & 100.0\% & 97.5\% & 100.0\% & 100.0\% & 97.5\% & 100.0\% \\
6 & qwen-2.5 & 32.5\% & 17.5\% & 92.5\% & 85.0\% & 95.0\% & 97.5\% & 100.0\% & 95.0\% \\
\hline
\end{tabular}
\caption{Accuracies for humans and LLMs across all variables. Reasoning models are bolded, and the human data is italicized. Human data was only collected for complexity 3, while LLM data spans complexities 3--6.}
\label{tab:complexity_results}
\end{table*}
\end{landscape}

\end{document}